\pdfoutput=1

\documentclass[11pt]{article}

\usepackage[]{naacl2024sub}
\usepackage{times}
\usepackage{latexsym}
\usepackage{multirow}
\usepackage[T1]{fontenc}

\usepackage[utf8]{inputenc}

\usepackage{microtype}

\usepackage{inconsolata}

\usepackage{float}

\usepackage[utf8]{inputenc} 
\usepackage{hyperref}       
\usepackage{url}            
\usepackage{booktabs}       
\usepackage{amsfonts}       
\usepackage{nicefrac}       
\usepackage{microtype}      
\usepackage{xcolor}         
\usepackage{xspace}
\usepackage{graphicx}
\usepackage{cleveref}
\usepackage{wrapfig}
\usepackage{subcaption}
\usepackage{bbding}
\newcommand{\thedataset}{EX-FEVER\xspace}

\usepackage{enumitem}
\setlist{leftmargin=*}
\setlist[itemize]{itemsep=3pt, topsep=3pt}
\usepackage{fontawesome5}
\usepackage{tcolorbox}

\title{\thedataset: A Dataset for Multi-hop Explainable Fact Verification}

\author{%
  Huanhuan Ma$^{1,2}$\quad Weizhi Xu$^{3}$\quad Yifan Wei$^{1}$\quad Liuji Chen$^{1,2}$\\ \quad \textbf{Liang Wang$^{1,2}$}\quad \textbf{Qiang Liu$^{1,2}$}
  \quad \textbf{Shu Wu$^{1,2}$}\thanks{\ \  To whom correspondence should be addressed.} \quad \textbf{Liang Wang$^{1,2}$}\quad \\
  $^{1}$ School of Artificial Intelligence, University of Chinese Academy of Sciences \\
  $^{2}$ New Laboratory of Pattern Recognition(NLPR)\\
  State Key Laboratory of Multimodal Artificial Intelligence Systems (MAIS)\\
  Institute of Automation, Chinese Academy of Sciences\\
  $^{3}$ ByteDance Inc. \\
  \normalsize\rule{0pt}{1em}
  \faEnvelope[regular]{} Primary contact: 
  \tt{huanhuan.ma@cripac.ia.ac.cn} 
}

\begin{document}

\maketitle

\begin{abstract}

Fact verification aims to automatically probe the veracity of a claim based on several pieces of evidence. Existing works are always engaging in accuracy improvement, let alone explainability, a critical capability of fact verification systems.
Constructing an explainable fact verification system in a complex multi-hop scenario is consistently impeded by the absence of a relevant, high-quality dataset.
Previous datasets either suffer from excessive simplification or fail to incorporate essential considerations for explainability.
To address this, we present \thedataset, a pioneering dataset for multi-hop explainable fact verification. With over 60,000 claims involving 2-hop and 3-hop reasoning, each is created by summarizing and modifying information from hyperlinked Wikipedia documents. Each instance is accompanied by a veracity label and an explanation that outlines the reasoning path supporting the veracity classification. 
Additionally, we demonstrate a novel baseline system on our \thedataset dataset, showcasing document retrieval, explanation generation, and claim verification, and validate the significance of our dataset.
Furthermore, we highlight the potential of utilizing Large Language Models in the fact verification task. 
We hope our dataset could make a significant contribution by providing ample opportunities to explore the integration of natural language explanations in the domain of fact verification.
\footnote{We make the \thedataset publicly accessible through \url{https://github.com/dependentsign/EX-FEVER}} 
\end{abstract}

\section{Introduction}
Fact verification, also known as fact checking, is a task to predict the veracity of a claim based on retrieved evidence, i.e., evidence that supports the claim, refutes the claim, or has insufficient information to judge the claim. Since the misinformation is widely spread with the proliferation of social platforms, recent years have witnessed the rapid development of automatic fact checking over various domains, such as politics~\cite{ostrowskiMultiHopFactChecking2021, wangLiarLiarPants2017, ContextAwareApproachDetecting2017}, public health~\cite{kotonya2020explainable, shahiFakeCovidMultilingualCrossdomain2020, nakovQCRICOVID19Disinformation2022}, and science~\cite{lazer2018science, wadden-etal-2020-fact}.


A typical fact-checking system consists of two main stages: evidence retrieval and veracity prediction. The evidence retrieval stage aims to improve the recall of golden evidence. The veracity prediction is made based on the interaction between the given claim and the retrieved evidence. 

The first large-scale fact-checking dataset, FEVER~\cite{thorne2018fever}, has significantly contributed to the promotion of existing works in the field
where claims are annotated by crowd workers and rely on information sourced from Wikipedia articles. The fact-checking systems inspired by FEVER aim to enhance both the performance (e.g., precision and recall) of evidence retrieval and the accuracy of verdict prediction~\cite{zhou2019gear, liu-etal-2020-fine}. However, it is important to acknowledge that over 87\% of the claims in FEVER rely on information sourced from a single Wikipedia article. In contrast, real-world claims often involve information from multiple sources, making it challenging for single-hop models to reason effectively without resorting to word-matching shortcuts~\cite{jiang2019avoiding}. To address this limitation, the HOVER dataset is introduced by~\citet{jiang_hover_2020}. The HOVER dataset is derived from the QA dataset~\cite{yang_hotpotqa_2018}, where claims require evidence from up to four English Wikipedia articles. 
The multi-hop design presents numerous challenges for both retrieval and verification models, but its oversimplification of the verification task as a binary classification is a major reason for its limited adoption in the fact verification task.

Though the research on multi-hop complex reasoning, the explainability is still under-explored. Explainability plays a significant role in fact-checking models for two main reasons. Firstly, displaying the veracity prediction along with the corresponding textual explanations can make the fact-checking system more credible to human users~\cite{atanasovaGeneratingFactChecking2020, kotonya2020explainable}. Moreover, the performance (e.g., F1 score, Accuracy) of models can be improved when the explanation is fed~\cite{stammbach_e-fever_nodate}. Hence, shedding light on the research of explainability in fact checking holds great value.

\begin{table}[h]
  \centering
  \begin{tabular}{llcc}
    \toprule
    \textbf{Datasets} & \textbf{Hops}  & \textbf{Explainable} & \textbf{Class} \\
    \midrule
    HOVER & 2-3-4     & \XSolidBrush & 2\\
    FEVER & 1-2      & \XSolidBrush  & 3\\
    e-FEVER & 1-2   & \checkmark  & 3 \\
    EX-FEVER & 2-3   & \checkmark  & 3 \\
    \bottomrule
  \end{tabular}
  \caption{Related Datasets Comparison}
  \label{tab:datasets}
\end{table}

Currently, only a few works in fact checking take textual explanations into consideration. The datasets LIAR-PLUS~\cite{alhindi2018your} and PUBHEALTH~\cite{kotonya2020explainable} involve about 10,000 claims with journalists' comments as explanations in politics and public health respectively, which are quite limited regarding the scale and the domain. Besides, though e-FEVER~\cite{stammbach_e-fever_nodate} complements textual explanations for the dataset FEVER using GPT-3, the quality of automatically generated explanations cannot be guaranteed. Furthermore, verifying the claims in all mentioned datasets always requires only one piece of information, i.e., one-hop fact-checking, where lots of methods have been proposed and achieved remarkable success~\cite{zhou2019gear, liu-etal-2020-fine}. However, generating precise explanations in complex, multi-hop fact-checking scenarios remains an open and challenging question.

\begin{figure*}[t]
   \begin{center}
   \includegraphics[width=0.85\linewidth]{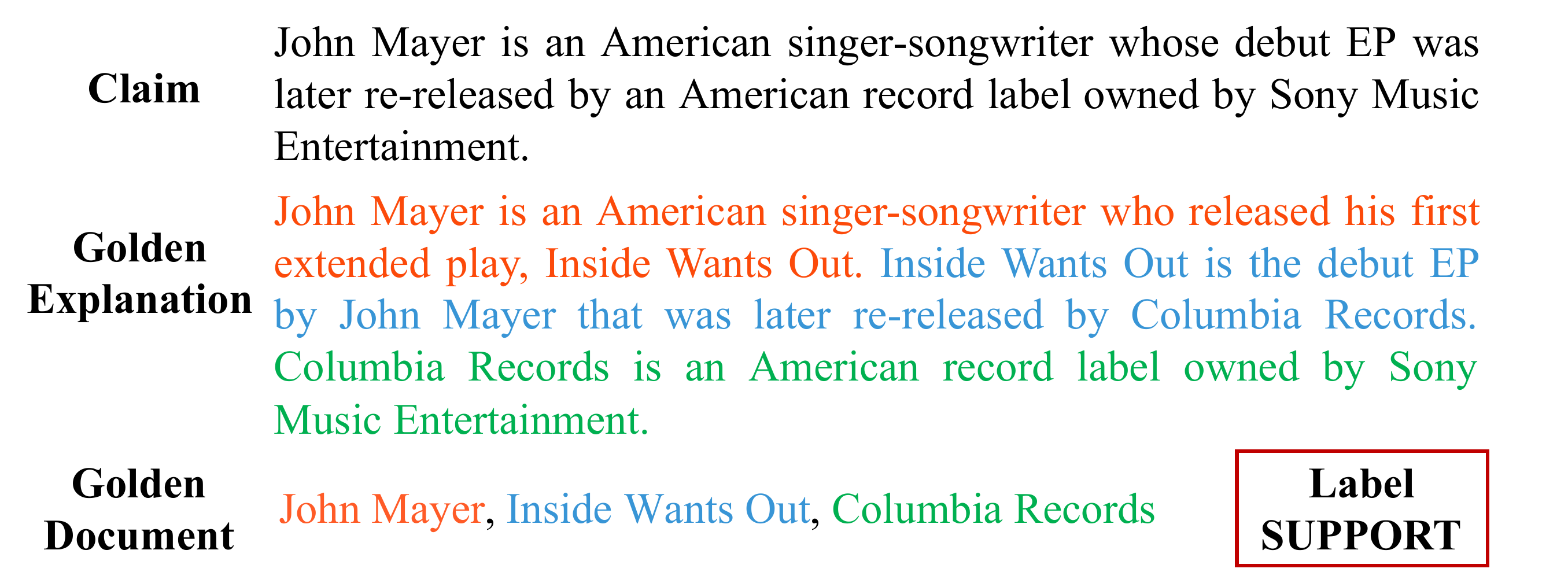}
   \end{center}
   \caption{A sample in the proposed dataset \thedataset. The textual explanation in different colors refers to the information in different documents.}
   \label{fig:example}
\end{figure*}

To facilitate the development of research around the aforementioned question, we propose the first dataset for multi-hop explainable fact verification, namely \thedataset. In general, the dataset involves 60,000 claims requiring 2-hop or 3-hop reasoning. Each claim is accompanied by two or three golden documents containing the necessary information for veracity reasoning. Different from existing datasets focusing on evaluating performance solely, a claim is assigned with not only a veracity label (SUPPORTS, REFUTES, and NOT ENOUGH INFO) but also a piece of golden explanation. The textual explanation describes the minimally sufficient information in each hop to verify a claim. Take Figure~\ref{fig:example} as an example, the veracity of the claim requires 3-hop reasoning, with each color representing the information from a different hop. Note that the explanation is possibly a segment of a complete sentence in the golden document and requires necessary rewriting since we want to keep the minimally sufficient information and other unrelated information will not be included, which is different from annotating a whole sentence as golden evidence in previous datasets. We employed crowd workers and more details can be seen in Section \ref{sec:data collection}.



Then we develop a baseline system and perform a comprehensive benchmark evaluation in Section~\ref{sec:baselines} which encompasses document retrieval, explanation generation, and verdict prediction. 
Through this evaluation, we achieve the following insights:
The effectiveness of the multi-hop design retrieval model~\cite{xiong_answering_2021}, and retrieval is a critical bottleneck in the fact-checking system.
The weakness of the graph-based verdict prediction model~\cite{zhou2019gear}.
The limitations of existing fact-checking models trained on previous datasets such as HOVER~\cite{jiang_hover_2020}. 
In Section~\ref{sec:prompt based}, we do some preliminary investigations using Large Language Models (LLMs) in two distinct ways: using LLMs as an actor and using LLMs as a planner finding that LLMs excel as a planner and generating explanations, rather than directly making predictions.

In a nutshell, our contributions can be listed as follows,
\begin{itemize}
    \item We propose the first dataset for multi-hop explainable fact verification, which can support and promote the development of such a challenging domain.
    \item We develop a baseline system to demonstrate the practical application of the dataset in various areas, including document retrieval, explanation generation, and verdict prediction. 
    \item We explore the potential of leveraging LLMs for fact verification in two distinct ways providing valuable insights into their effectiveness and applicability in this context. 
\end{itemize}

\section{Related Work}

In this section, we briefly review representative works in the field of fact verification and model explainability.

\subsection{Fact Verification}
Fact verification is a task similar to natural language inference, where the target is to predict whether evidence entails a claim. 
There are several models carefully designed for veracity reasoning based on retrieved evidence, including transformer-based methods \cite{jiang2021exploring, kruengkrai2021multi} and graph-based methods~\cite{zhou2019gear,liu-etal-2020-fine,zhong2020reasoning,xu-etal-2023-counterfactual,weizhiwww,gong2024heterogeneous}. 
Existing works can be roughly grouped into two categories aiming to enhance performance and robustness, respectively.

Though precision-targeted models achieve satisfactory performance, some researchers discovered the precision improvement is derived from dataset biases, where these models would suffer from significant performance decline under different data distributions, i.e., the poor robustness~\cite{Schuster2019TowardsDF}. To this end, several unbiased datasets Symmetric~\cite{Schuster2019TowardsDF}, FEVER2.0~\cite{thorne-etal-2019-evaluating}, and FM2~\cite{eisenschlos-etal-2021-fool} are proposed to evaluate the model robustness. 

While the research on precision and robustness achieved remarkable results, they were not of explainability. 
Different from the former two problems, this paper focuses on the explainability of the fact-checking system, which still lacks a fundamental benchmark, including both a dataset and an evaluation system.  

\subsection{Model Explainability}
The generation of natural language explanations is treated as one of the ways to reveal the mechanism inside the `black-box' deep learning model~\cite{Ribeiro0G16,camburu2018snli,JollyAA22}. 
Generally, there are two pipelines, i.e., extractive approaches~\cite{yang_hotpotqa_2018,atanasovaGeneratingFactChecking2020,inoue-etal-2021-summarize} and abstractive approaches~\cite{kotonya2020explainable}. 

In real-life applications, the extracted approaches may contain redundant information and pronouns lacking context, which is not ideal. Therefore, the abstractive approach, which aims to provide concise and contextually understandable explanations by filling in the pronouns with an understanding of the context, has become a more popular and reasonable method for achieving explainability.

Despite its importance and growing research interests, there are only a few datasets in fact verification considered non-extracted approaches.
Fact-checking datasets LIAR-PLUS~\cite{alhindi2018your} and PUBHEALTH~\cite{kotonya2020explainable} involve about 10,000 claims with journalists’ comments as the explanation in politics and public health respectively, which is quite limited regarding the scale and the domain. 
Although \citet{stammbach_e-fever_nodate} made efforts to generate textual explanations for the FEVER dataset using GPT-3, automatic machine-generated interpretation text cannot guarantee quality.
In contrast, our \thedataset offers a comprehensive platform on a large scale, offering high-quality human-annotated explanations in the intricate multi-hop scenario.

\section{Data Collection}
\label{sec:data collection}
In this section, we provide a detailed description of the specific steps involved in annotating claims, including their corresponding labels and explanations.

\subsection{Claim Annotation}
Given a set of evidence documents, we initially require annotators to write the SUPPORTS claim first, and then create the corresponding REFUTES claim and NOT ENOUGH INFO claim by modifying the SUPPORTS claim. By doing so, we obtain three distinct claims, each labeled differently. These claims are based on a set of 2 or 3 golden documents, where the number of documents corresponds to the information hops involved in the claims. For brevity, we refer to these claims as \textbf{SUPPORTS}, \textbf{REFUTES}, and \textbf{NOT ENOUGH INFO} claims.

\textbf{SUPPORTS Claim Creation.}
In this class, claims must incorporate information from related golden documents, which are interconnected via hyperlinks. Annotators replace phrases in the initial document with their corresponding descriptions from hyperlinked documents, creating a 2-hop claim. Similarly, if there are three documents, this process is iterated to form a 3-hop claim.
For instance, consider the following example: the Wikipedia article for "Bohemian Rhapsody" states that \emph{Bohemian Rhapsody" is a song by the British rock band Oucen. It was written by Freddie Mercury for the band's 1975 album  Night at the Opera}. Correspondingly, the Wikipedia page for "A Night at the Opera" further elucidates that \emph{A Night at the Opera is the fourth studio album by the British rock band Queen, released on 21 November 1975 by \ldots}. In this context, the generated claim would read as follows: \emph{"Bohemian Rhapsody" is a song by the British rock band Queen for their1975 music album that was released on 21 November 1975.}

\textbf{REFUTES Claim Creation.}
To generate contradictory REFUTES claims, we follow a process of modifying SUPPORTS claims. Similar to previous fact-checking datasets \citep{jiang_hover_2020}, we utilize several mutation methods to ensure dataset diversity. These methods include:

\begin{itemize}
\item \emph{Entity Replacement:} Entities in a claim play a crucial role in determining its veracity. By substituting the entity in a SUPPORTS claim with an unrelated entity, we can transform it into a REFUTES claim. This change creates a semantic conflict with the evidence documents. Common entities that are frequently replaced include names, places, and organizations. For instance, consider the original claim: \emph{Love \& Mercy was about the leader of an Australian rock band consisting of an American musician, singer, and songwriter.} By changing the country from American to Australian, the claim now conflicts with the evidence documents.

\item \emph{Logical Word Replacement:} This method tests the fact-checking model's ability to reason logically. It involves replacing logical words, such as comparative adjectives or temporal phrases, to alter the semantics in the opposite direction. For example, the word `smaller' can be replaced with `larger' to reverse the meaning. Similarly, temporal phrases like "during the 1960s" can be modified to "before the 1960s" or "after the 1960s." Additionally, specific time periods can be replaced with other relevant time frames.

\item \emph{Negation Word Insertion/Removal:} This method aims to assess the model's comprehension of negations within a sentence. Annotators are instructed to remove negation words, such as `not' and `never,' from SUPPORTS claims if they exist. Alternatively, negation words can be inserted into the SUPPORTS claim, or adjectives can be substituted with their antonyms. It's important to note that while it is necessary to include some REFUTES claims using this method, caution must be exercised. Some researchers have observed that models tend to establish spurious relationships between negations and the `REFUTES' label, leading to biased and unreliable predictions. Therefore, annotators are encouraged to primarily utilize the first two methods, resorting to this method only if necessary. This approach helps maintain dataset balance and mitigate biases.
\end{itemize}

\textbf{NOT ENOUGH INFO Claim Creation.}
There are two methods to create the claim in this class from both the evidence side and the claim side. Firstly, we randomly remove one of the shown documents to create a lack of evidence information. Then, the NOT ENOUGH INFO claim is the same as the SUPPORTS claim. The second method is to write a claim unrelated to the evidence information and keep the documents unchanged.

For 2-hop claims, we all utilize the second method since there are only two documents, and removing one of them will leave only one piece of information, which is easy for models to detect. For 3-hop claims, we randomly employ two methods to generate NOT ENOUGH INFO claims with the same probability. 

\subsection{Explanation Annotation}
After writing claims, we ask annotators to write textual explanations that demonstrate how to reach the veracity label based on the evidence.

\textbf{The Explanation for SUPPORTS Claim} contains minimally sufficient information to draw a supported conclusion. Specifically, there are two (three) sentences for a 2-hop (3-hop) claim in explanation, where each sentence is the summarization of claim-related information in each golden document. Such an explanation describes the reasoning path among the given documents. 

\textbf{The Explanation for REFUTES Claim.}
Different from the explanation in the class SUPPORT, it not only involves the summarization of each golden document, but also points out which part of the claim is inconsistent with the documents. For instance, the claim is that \emph{"Bohemian Rhapsody" is a song by the British rock band Queen for their 1975 music album that was released on 21 December 1975}, the corresponding explanation would read as follows: \emph{"Bohemian Rhapsody" is a song by the British rock band Queen from their 1975 album A Night at the Opera. A Night at the Opera was released on 21 November 1975, not 21 December 1975}. The explanation not only describes the reasoning process but also pinpoints the specific parts of the claim that are inconsistent with the relevant golden documents.

\textbf{The Explanation for NOT ENOUGH INFO Claim.}
Since there are two types of NOT ENOUGH INFO claims generated via different methods, the corresponding explanation is also distinct. In detail, for claims whose evidence documents are randomly removed, we ask annotators to indicate which piece of the necessary information is dropped and involve information in the preserved documents in the explanation. For claims unrelated to the evidence document, the explanation should be \emph{there is no information to verify the claim.} For example, the claim is \emph{Louise Simonson won an award for Outstanding Achievement in Comic Arts}, the corresponding explanation will be \emph{There is no information showing that Louise Simonson won an award for Outstanding Achievement in Comic Arts}.

\subsection{Overall Annotating Process}
We employed the annotators from Appen\footnote{\url{https://www.appen.com}} and supplied them with our detailed annotation guidelines. 
We first identify a seed article from the top 50,000 popular Wikipedia pages to initiate the process.
Then we spend two weeks training those annotators from our feedback after the quality is consistent and high enough.




To ensure the collection of a high-quality dataset, we engage additional annotators as quality inspectors to meticulously review the annotations submitted by the primary annotators. Each piece of collected data undergoes consistency checks by the annotators to guarantee quality.
Further details can be found in \ref{sec:app:datacollection}.

\subsection{Data characteristic}
In summary, we collected over 60,000 claims featuring three distinct labels while maintaining label balance, with each label constituting approximately 33\% of the dataset. The label and hop count distribution is described in table~\ref{tab:datastatistic}. We divide the dataset into training, validation, and test subsets based on a 70\%-20\%-10\% split. The claim mean length and the explanation length at the word-level are shown in table \ref{tab:datastatistic}.
We also randomly extract a mini-test data subset from the full test dataset, comprising 1,000 claims, and utilize this to test the capabilities of large language models (e.g., ChatGPT).

\begin{table}[h]
\centering
\caption{Data Statistics with different number of hops and different label classes. The average claim length and explanation length in word level are reported}
\label{tab:datastatistic}
\resizebox{0.50\textwidth}{!}{
\begin{tabular}{llcccc}
\toprule
\textbf{Hops} & \textbf{SUP} & \textbf{REF} & \textbf{NEI} & \textbf{Claim} & \textbf{EXP} \\
\midrule
2 Hops & 11053 & 11059 & 11412 & 21.63 & 28.39 \\
3 Hops & 9337 & 9463 & 8941 & 30.69 & 43.45 \\
Total & 20390 & 20522 & 20353 & 25.73 & 35.21 \\
\bottomrule
\end{tabular}
}
\end{table}

\begin{figure*}[!h]
   \begin{center}
   \includegraphics[width=0.90\linewidth]{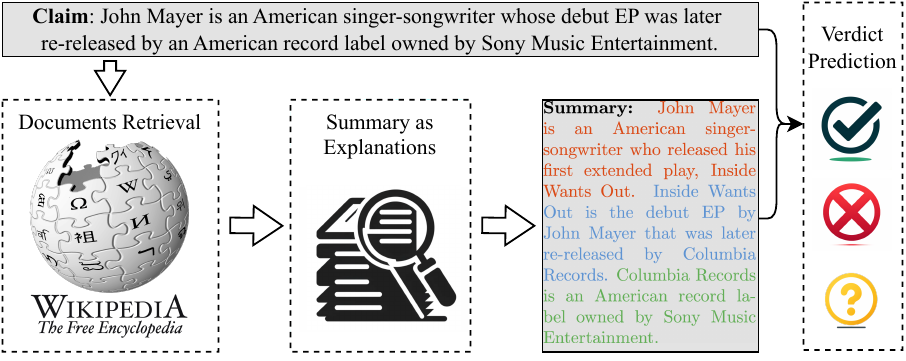}
   \end{center}
   \caption{The baseline system comprises three stages: document retrieval, summary generation as explanations, and verdict prediction. The system produces two main outputs: a veracity label indicating whether the claim is 'SUPPORT'ed, 'REFUTE'd, or there is 'NOT ENOUGH INFO', and a summary that serves as an explanation for the prediction.}
   \label{fig:example}
\end{figure*}

\section{Baseline System Description}
\label{sec:baselines}

Our baseline system comprises three stages. Firstly, given a claim, the system aims to provide the most relevant documents as evidence. Secondly, the system summarizes the information from multiple documents into a concise summary, which then serves as the system's output explanation. Finally, based on the interaction between the claim and the summary, the system generates a verdict.
In the subsequent sections, we will delve into the specific models employed in each stage.

\textbf{Document Retrieval}
At this stage, given a claim $c$, the objective is to retrieve relevant documents $D = d_1, \cdots, d_N$. Firstly, we employ a rule-based document retrieval based on the DrQA system~\cite{chen-etal-2017-reading}, using cosine similarity between binned unigram and bigram Term FrequencyInverse Document Frequency (TF-IDF) vectors. Then we use two neural-based document retrieval models, the BERT-based model~\cite{kenton2019bert} and the MDR model~\cite{xiong_answering_2021}.
The BERT-based model calculates the similarity score between each document $d$ and the given claim $c$. 
The MDR model incorporates a multi-hop retrieval design, which iteratively selects a document based on the probability modeled through a query reformulation process conditioned on pre-retrieval results. 
Finally, we select the top-5 documents feeding into the next stage.

\textbf{Explanatory Stage}
This stage aims to use the model to understand the relevant documents and produce a piece of minimally sufficient summarization as the system explanation. 
We fine-tune a BART~\cite{lewis-etal-2020-bart} model, which is the state-of-the-art model in text generation and summarization tasks, with the given supervised golden explanation on our dataset's training split.

\textbf{Verdict Predict}
In this stage, the system needs to make a verdict based on the interaction between the claim and the generated summary explanation.
We adapt a transformer-based method BERT\cite{kenton2019bert}.
As a comparative alternative, we implement a graph-based text reasoning model, the state-of-art fact-checking model GEAR~\cite{zhou2019gear}.

\textbf{Aug-HOVER}
To further evaluate the importance of our dataset, we conduct training on the previous dataset HOVER~\cite{jiang_hover_2020}, but the system is tested on our own dataset.
In this approach, the \thedataset data is not used for training the verification model. Instead, it is exclusively utilized for validation and testing purposes. The verification model is trained solely on the Hover dataset. Since the Hover dataset consists of binary class labels, we merge the "NOT ENOUGH INFO" class and the "REFUTE" class from \thedataset, creating a unified "NOT\_SUPPORTED" class for evaluation. The remaining data is aligned with the Hover dataset.

\subsection{Experimental results}

\begin{table}[h]
\caption{Retrieve Model Performance Comparison}
\label{tab:retrieve_model_comparison}
\centering
\resizebox{0.49\textwidth}{!}{
\begin{tabular}{lcccc}
\toprule
\textbf{Model}  & \textbf{EM} & \textbf{Hit@6} & \textbf{Hit@12} & \textbf{Hit@30} \\
\midrule
MDR  & 43.3 & 55.00 & 60.90 & 68.60 \\
BERT-based & 32.4  & 66.12 & 70.28 & 73.98 \\
\bottomrule
\end{tabular}
}
\end{table}

\begin{table*}[h]
\caption{Generated Summary Metrics Comparison}
\label{tab:summary_metrics}
\centering
\resizebox{0.75\textwidth}{!}{
\begin{tabular}{lcccccc}
\toprule
\textbf{Model}  & \textbf{Length} & \textbf{rouge1} & \textbf{rouge2} & \textbf{rougeL} & \textbf{rougeLsum} \\
\midrule
MDR & 54.79 & 54.88 & 41.34 & 49.42 & 53.02 \\
BERT-based & 46.05 & 46.88 & 32.80 & 35.52 & 44.41 \\
\midrule
Explanation from ChatGPT & & & & & \\
\midrule
GPT-0example & 58.05 & 52.28 & 33.74 & 48.13 & 49.89 \\
GPT-3example & 48.56 & 59.98 & 42.85 & 57.66 & 55.61 \\
\bottomrule
\end{tabular}
}
\end{table*}

\begin{table*}[h]
\caption{Verify Model Comparison. The accuracy (\%) of each model is reported}
\label{tab:verify_model_comparison}
\centering
\resizebox{0.75\textwidth}{!}{
\begin{tabular}{lllcc}
\toprule
\textbf{Model} & \textbf{Val}  & \textbf{Test}& \textbf{Test On Golden} & \textbf{Train With Golden}\\
\midrule
Gear@BERT-based &  54.96  & 54.71  & 53.08    & 61.05        \\
Gear@MDR & 59.68    & 58.89  & 53.98       & -        \\
BERT@BERT-based & 68.07    & 67.65  & 76.69  &  99.29  \\
BERT@MDR & 73.86    & 73.34  & 76.89  &  -  \\
HOVER@MDR & 46.58  & 45.41  & 33.79  & - \\
\bottomrule
\end{tabular}
}
\end{table*}

\textbf{Document Retrieval}
Our evaluation metrics of choice are the exact match (EM) score and the hit score. The experimental results are presented in Table \ref{tab:retrieve_model_comparison}, demonstrating that the MDR model achieves a better EM score, while the BERT-based model obtains a higher hit score. \textbf{It is possible that in scenarios involving multiple hops, iterative information retrieval (MDR) is effective in achieving a better EM score.}
\textbf{Additionally, the hit score improves marginally with an increasing number of hits.}

\textbf{Explanatory Stage}
We choose the Rouge score as an evaluation metric to assess the quality of the generated explanation which measures the similarity between the golden explanation and the generated explanation.
The results are summarized in Table \ref{tab:summary_metrics}.
When fed into the documents retrieved from the MDR model, BART achieves a superior rouge score, revealing that \textbf{the retrieval model is the bottleneck of the system.}
Moreover, this observation further validates the importance of the EM score, as we cannot simply increase the number of hits due to the input length constraint of the text-generating model.

\textbf{Verdict Predict}
Table \ref{tab:verify_model_comparison} displays the final verdict outcomes. The BERT model outperforms the graph-based method GEAR in terms of accuracy, whether using the MDR retrieval model or the BERT-based retrieval model, by a significant margin. To further evaluate the models, we conducted tests using golden explanations, which represent an ideal scenario assuming a perfect explanation-generating model. Surprisingly, the performance of the GEAR model did not improve. \textbf{This finding suggests that} \textbf{the graph method relies more on pattern recognition rather than genuine reasoning capabilities.} When the input data shifted from generated explanations to golden explanations, the GEAR model failed to show any improvement.

Furthermore, we devised an approach for training the verdict prediction model using golden explanations. Although this setting is not practical for real-world usage, it allows us to assess the quality of our golden explanations. In this scenario, the verdict prediction model achieved optimum accuracy, indicating the high quality of our golden explanations.

\textbf{Aug-HOVER}
The experimental results of Aug-HOVER are also showcased in Table \ref{tab:verify_model_comparison}.
The obtained test results demonstrate a performance of only approximately 45\%, indicating a subpar outcome.\textbf{ This suggests that the verification model trained on the previous dataset may not be capable of addressing the task we proposed which further validates the importance of our dataset.}

\section{Prompt based approach}

\label{sec:prompt based}

With the PLMs (pre-trained language models) and GPT model series development~\cite{brown_language_2020, radford_improving_nodate, radford_language_nodate, ouyang_training_2022}, large language models (LLMs) exhibit immense potential in many general tasks, especially with suitable prompts~\cite{NEURIPS2022_9d560961, wei2023menatqa}. 
In this section, we do some preliminary investigations exploring using LLMs in the fact checking task in two directions: directly using LLMs as an actor, and using LLMs as a planner also we both evaluate the verdict accuracy and the ability of LLMs to generate explanations.

\subsection{LLMs as an actor}
In this approach, we directly prompt an LLM as an actor instructing ChatGPT to directly make a verdict with a given claim. 
We evaluate both its ability in verdict prediction and explanation generation. We use a variety of prompt templates which include only giving the claim, giving the claim with the golden documents, adding the few shot examples, giving the instruction in json format, and instructing GhatGPT with or without requiring explanation, and details are shown in~\ref{sec:appendix}. We also extract the explanation from LLMs's responses to evaluate the explanation quality.

\subsection{LLMs as a planner}
In this approach, we adopt the methodology outlined by~\cite{pan2023factchecking}. Following their setting, we instruct ChatGPT to generate program guides for each individual claim. Subsequently, we employ a verification model to execute these program guides, thereby obtaining the verdict prediction for each claim.

\subsection{Results dicussion}

\begin{table}[h]
\caption{Use LLM as an actor or a planner. The accuracy (\%) of each model is reported.}
\label{tab:program}
\centering
\begin{tabular}{llccc}
\toprule
\textbf{Type} & \textbf{Model}  & \textbf{Close}& \textbf{Open} & \textbf{Gold}\\
\midrule
\multirow{5}{*}{Actor} & ClaimOnly & 45.78 & - & - \\
 & w/o exp & - & - & 47.91 \\
 & w/ exp & - & - & 47.92 \\
 & 1 shot & - & - & 47.91 \\
 & 3 shots & - & - & 58.69 \\
\midrule
Planner & ProgramFc &  47.30   & 51.70  & 64.90      \\
\bottomrule
\end{tabular}
\end{table}

The verdict accuracy of the two paradigms is reported in table~\ref{tab:program}, and to evaluate the quality of the explanation generated from ChatGPT, we use Rouge score, and the test result is appended in table~\ref{tab:summary_metrics}.

In the context of employing the "LLMs as an actor" paradigm, when we provide only the claim without relevant evidence documents, necessitating LLMs to rely on their internal knowledge for predictions, ChatGPT yields the lowest favorable result with an accuracy of 45.78\%. \textbf{This outcome may suggest that despite the vast training data LLMs have been exposed to, they still need extra knowledge to perform this task.}
When we introduce a few shot examples to assist LLMs, a noticeable enhancement in performance becomes apparent. Furthermore, as the quantity of provided examples increases, there is a corresponding improvement in performance, from 47.91\% to 58.68\%. \textbf{This suggests that the incorporation of few-shot, in-context learning proves to be an effective approach for addressing the task at hand in our study.}

Nonetheless, within the "LLMs as a planner" paradigm, we do not directly obtain verdict results from ChatGPT itself. Instead, we solely rely on the large model to generate program guides, while the verification model employed is a non-finetuned 6-billion-parameter FLan-T5 model\cite{FLan-T5}. Surprisingly, this approach yields substantial improvements in performance.

In contrast to the "LLMs as an actor" approach, when provided with golden evidence, the accuracy increases from 58\% to 64\%. \textbf{This intriguing phenomenon appears to suggest that the large model excels not in making predictions but rather in serving as a planner, generating guides to facilitate judgments by other models.}

Then, we conducted a test using ChatGPT to generate explanations for after making the veracity predictions. The results of this test are appended in Table~\ref{tab:summary_metrics}.
We observe that ChatGPT's performance in explanation generation improves as the number of input examples increases. When we provide three examples, ChatGPT outperforms the fine-tuned Bart model.
\textbf{That suggests that ChatGPT performs better in the explanation generation task compared to making predictions for claims.}

\section{Conclusion}

We introduce a publicly accessible, extensive fact-checking dataset, named \thedataset, encompassing over 60,000 intricate multi-hop claims.
For each claim's veracity, we provide an elucidating annotation to facilitate human comprehension of the adjudication process. 
We devise a comprehensive system using \thedataset, encompassing a retrieval phase, a summarization component serving as the explanatory stage, and a subsequent verification stage. 
The experiment results validate the challenge and the importance of \thedataset.
Furthermore, we do some preliminary investigations exploring using LLMs in the fact checking task.
We elect the GPT-3.5-turbo model to represent large language models. 
Utilizing two distinct settings using LLMs as an actor and using LLMs as a planner, we evaluated the LLM on our dataset mini-test set. 
We find that the LLM exhibits better performance when utilized as planners rather than directly employed as actors and that LLM excels in generating explanations rather than making predictions.
The findings reveal that despite the prowess of LLMs, there remains significant potential for improvement.
In summary, \thedataset could serve as a valuable benchmark in studying the explainable multi-hop fact-checking task by improving reliability, trustworthiness, and facilitating better decision-making across different domains.

\section*{Limitations}
Although the intention behind proposing this dataset is to bridge the gap between human experts and automated fact verification systems in terms of explainability, evaluating the effectiveness of the system remains challenging. In this paper, we utilize the Rouge score to assess the system's output explanation quality. However, to evaluate the output explanation quality the Rouge score is insufficient. There are instances where different lexical choices in the explanation could yield the same result, but the Rouge score, which is based on similarity, may penalize such cases, leading to inaccurate evaluation results.
\textbf{Hence, further investigation is warranted to explore an evaluation scoring mechanism that goes beyond mere similarity and effectively assesses the quality of explanations.}

Another limitation is that in the retrieval stage, a situation arises where information can be sourced from more than one Wikipedia document. While we have deliberately simplified and overlooked this scenario to enhance the manageability of the annotation process, it's important to acknowledge that models might receive undue penalties when retrieving information from these sources, particularly when the retrieved documents aren't in the golden documents.

\section*{Ethics Statement}

\paragraph{Biases.}
Our data is collected based on the top 50,000 popular Wikipedia pages. Our process does not introduce any additional biases, although there may be inherent biases present in Wikipedia that are beyond our control.

\paragraph{Intended Use and Misuse Potential.}
Our dataset can be valuable for enhancing the development of auto-fact-checking systems in areas such as document retrieval, document summaries, and verdict prediction. To the best of our knowledge, there is no specific potential for misuse associated with this dataset, or at least no more potential for misuse than other fact verification datasets.

\section*{Acknowledgement}

This work is supported by National Natural Science Foundation of China (62372454,  62141608)

\bibliography{naacl2024sub}

\appendix
\section{Appendix}
\label{sec:appendix}

\subsection{Details of data collection}
\label{sec:app:datacollection}
\begin{figure*}[h]
       \begin{center}
       \includegraphics[width=0.99\linewidth]{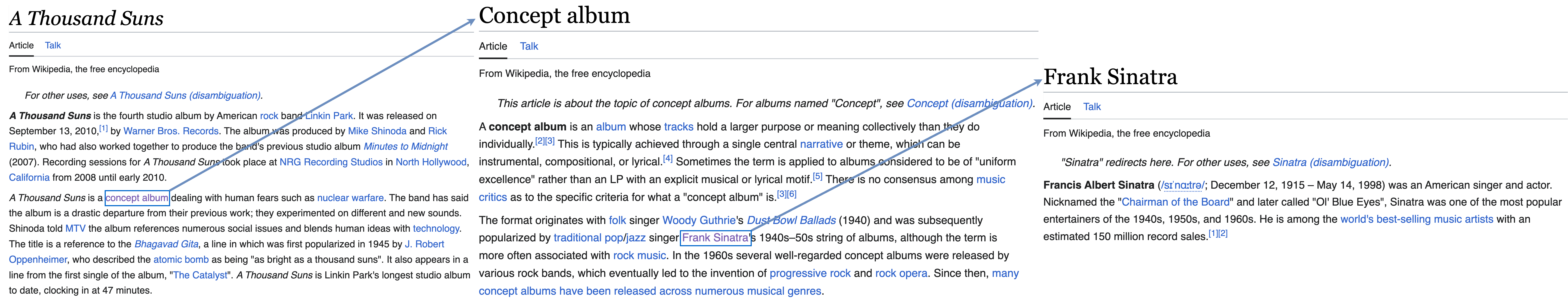}
       \end{center}
       \caption{A sample in the proposed dataset \thedataset. The corresponding claim is "A Thousand Suns is an album dealing with human fears such as nuclear warfare, where the theme of the album was subsequently popularized by a traditional pop/jazz American singer and actor" }
       \label{fig:example_collection}
\end{figure*}

\begin{figure*}[h]
  \centering
  \includegraphics[width=0.98\textwidth]{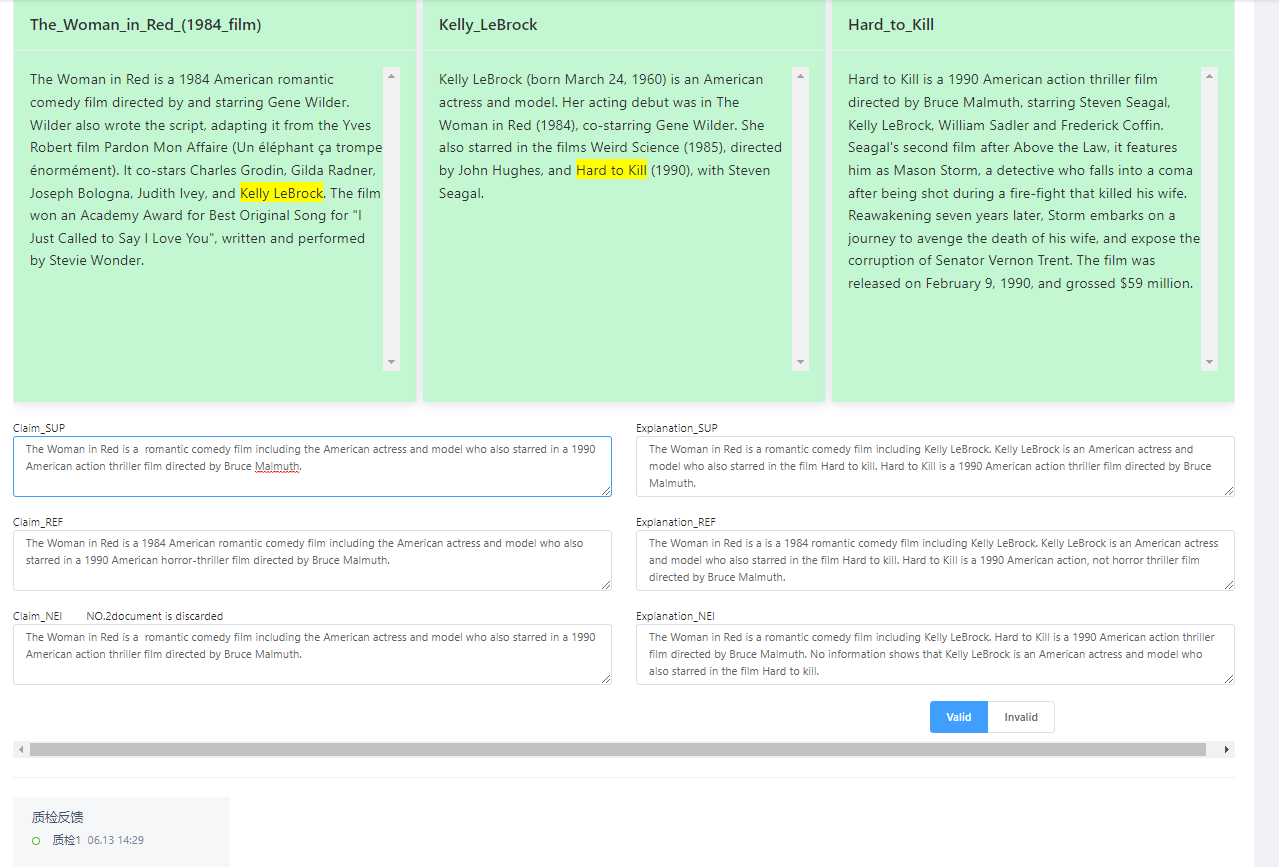}
  \caption{Annotation platform}
  \label{fig:annotation_platform}
\end{figure*}

We employ annotators from Appen and supply them with our detailed annotation guidelines. To initiate the process, we first identify a seed article from the top 50,000 popular Wikipedia pages and then select subsequent articles based on the hyperlink entities on the first page. We iterate this process one more time to get 25 instance pairs, which consist of three Wikipedia entities as the multi-hop reasoning path. Then we randomly sample some pairs from the 25 pairs to avoid position bias. Since the bridge entity might not be reasonable enough, the annotator is able to discard this instance to keep more reasonable pairs, and the skipped data is discarded.

Following this, annotators proceed with the annotation process as previously described. We first spend two weeks training those annotators from our feedback, after the quality is consistent and high enough.

We first ask annotators to read two or three Wikipedia documents that are highlighted to feature a linked entity. From there, they generate a claim that encompasses all the golden documents labeled as SUPPORT and construct REFUTE examples. For the NOT ENOUGH INFO label, we employ two approaches: either randomly removing an article or composing a claim that is unrelated to the current article information. Upon completion of these steps, the submitted data is handed over to quality inspectors for further examination.

In order to harvest a high-quality dataset, we employ additional annotators as quality inspectors to scrutinize the annotations submitted by the primary annotators. If the data meets the inspection criteria, it is accepted; however, if the data is found to be inadequate, it is returned to the annotators along with revision suggestions for re-annotation until it successfully passes inspection. We ask annotators to annotate every 15,000 instances as a milestone. Then we roll a data validation period.

We collect Wikipedia article pairs and then engage annotators from Appen to annotate a total of 60,000 claims. Each claim is assigned a verdict label and an accompanying explanation. In total, we compensate the crowdworkers with a sum of 130,000 CNY. Each instance costs about 2.2 CNY. These crowdworkers are hired from Appen who are fluent in English.


\begin{table*}[h]
  \centering
  \begin{tabular}{lccccc}
    \toprule
    \textbf{Datasets} & \textbf{Multi-hop} & \textbf{Statement Count} & \textbf{Explanation Type} & \textbf{Class Count} & \textbf{Source} \\
    \midrule
    HOVER & \checkmark   &  26,171   & None & 2 & Wiki\\
    FEVER & \XSolidBrush & 185,445  & None  & 3 & Wiki\\
    e-FEVER & \XSolidBrush  & - & Machine Generated  & 3 & Wiki\\
    EX-FEVER & \checkmark   & 60,000  & Human Annotated  & 3 & Wiki\\
    LIAR-PLUS & - & 12,836 & Extracted Justification & 6 & Fact Check\\
    PUBHEALTH & - & 11,832 & Justification & 4 & Fact Check\\
    \bottomrule
  \end{tabular}
  \caption{Comparison of Related Datasets. Note: e-FEVER dataset size is not revealed. LIAR-PLUS and PUBHEALTH are crawled from fact-check websites and are not practical for multi-hop classification}
  \label{tab:datasets}
\end{table*}

\subsection{Details of the baseline system implementation}
\textbf{Document Retrieval}
Given a specific claim $c$, we initially utilize the TF-IDF model to generate the top 200 documents, denoted as $D = d_1, \cdots, d_N$, where $N = 200$. Subsequently, both neural-based retrieval models access the resulting document corpus. The two neural-based models are fine-tuned on our dataset's training split.

For the BERT-based mode, during the training phase, we select five documents for each claim. This training set comprises two (three) golden documents and three (two) non-golden documents that possess the highest cosine similarity scores from the TF-IDF model.  
The BERT-based model is fine-tuned on positive-negative pairs data from the dataset. This model takes a single document $d \in D$ and the claim $c$ as inputs, and outputs a score that reflects the relatedness between $d$ and $c$. 
During the testing phase, the model evaluates the relevance score for each claim-document pair and then sorts the documents $D$ according to their relevance score.

The MDR model uses a shared RoBERTa-base~\cite{liu2019roberta} encoder for both document and query encoders and implements maximum inner product search over dense representations of the pre-retrieval documents set.

The process of MDR is slightly different. For each claim, we select either the first or second document with the highest ranking from the TF-IDF model as the hard negative sample. Since the retrieval process of MDR operates in an iterative manner, with each query relying on the results from the previous retrieval, we conduct experiments differentiating between 2-hop and 3-hop data. In a single training epoch, we randomly assign MDR to initially train on either the 2-hop or 3-hop data. During the testing phase, we retrieve three documents for each claim. However, when calculating the evaluation metrics, we disregard the third document for the 2-hop portion of the data. 

\textbf{Explanatory Stage}
We fine-tune the BART model via the Hugging Face Transformers library\footnote{\url{https://github.com/huggingface/transformers/tree/main/examples/pytorch/text-classification}\label{huggingface}} on our dataset. To enable the model to capture all relevant information during training, all golden documents are included in the model's input. Similar to the Document Retrieval experimental setting and given that the BART model has limited input tokens, the training data comprise five documents for each record, two (three) golden documents and three (two) non-golden documents that possess the high-rank order from the retrieval model. In contrast, during the test phase, we select the top five documents yielded by the retrieval models without additional consideration for including all golden documents and we select the rouge score as the evaluation metric. 

\textbf{Verdict Predict}
We designate the claim to be verified as the MNLI task hypothesis and the relevant explanation of the verdict as the MNLI task premise. 
The BERT model is fine-tuned by using the Hugging Face Transformers library\footref{huggingface}.
We implement the GEAR model~\cite{zhou2019gear}, a graph-based text reasoning model, as a comparative alternative. The GEAR model conceptualizes sentences functioning either as evidence or claims within a graph theoretic framework, regarding such sentences as nodes on a graph. An evidential reasoning network and evidential aggregator transmit evidential information and make predictions. By combining these elements, GEAR leverages both evidence and claim in BERT to obtain an evidence representation $e_i$. The claim is fed into BERT alone to obtain representation $c$. Translating evidential and claim sentences into graph nodes expressing their relations, GEAR provides an integrative approach to adjudicating claim veracity via reasoning across the evidence.

\subsection{Details of GPT-3.5-turbo Prompt}

\begin{tcolorbox}[colback=gray!5!white,colframe=gray!75!black,title=Claim only]
Check the claim: [claim] \\
Choices:['SUPPORT', 'REFUTE','NOT ENOUGH INFO']\\
Answer: 
\end{tcolorbox}

\begin{tcolorbox}[colback=gray!5!white,colframe=gray!75!black,title=W/o explanation required]
Claim: [claim]\\
Evidence: [golden documents]\\
Evaluate the claim based on the provided evidence and\\
choose one of the following labels: 'SUPPORT', 'REFUTE', or 'NOT ENOUGH INFO'.
\end{tcolorbox}


\begin{tcolorbox}[colback=gray!5!white,colframe=gray!75!black,title=W/ explanation required]
Claim: [claim]\\
Evidence: [evidence]\\
Evaluate the claim based on the provided evidence and\\
choose one of the following labels: 'SUPPORT', 'REFUTE', or 'NOT ENOUGH INFO'.\\
Provide a brief explanation for your choice.
\end{tcolorbox}

\begin{tcolorbox}[colback=gray!5!white,colframe=gray!75!black,title=Few-shot prompt]
I will provide you with evidence and a claim. Your task is to determine if the claim is supported, refuted, or if there is not enough information based on the given evidence. You need to choose one of the following labels: 'SUPPORT', 'REFUTE', or 'NOT ENOUGH INFO'. After choosing a label, please provide a brief explanation for your choice.

Example 1:\\
Evidence: [evidence]\\
Claim: [claim]\\
Choices: ['SUPPORT', 'REFUTE', 'NOT ENOUGH INFO']\\
Answer: [answer]\\
Explanation: [explanation]\\
...\\
Example N:\\
Evidence: [evidence]\\
Claim: [claim]\\
Choices: ['SUPPORT', 'REFUTE', 'NOT ENOUGH INFO']\\
Answer: [answer]\\
Explanation: [explanation]\\
Now, please evaluate the following claim based on the provided evidence:\\

Evidence: [evidence]\\
Check the claim: + [claim] + from the above evidence\\
Choices: ['SUPPORT', 'REFUTE', 'NOT ENOUGH INFO']\\
Answer:
\end{tcolorbox}
where N = 1, or 3.

\end{document}